\documentclass[]{arxiv_article}

\usepackage[toc,page,header]{appendix}

\microtypesetup{expansion=false}
\usepackage{xspace}
\usepackage{enumitem}
\usepackage[linesnumbered,ruled,vlined]{algorithm2e}
\usepackage{amsmath}
\usepackage{amssymb}
\usepackage{multirow}
\usepackage{booktabs}
\usepackage{graphicx}
\usepackage{pifont}

\definecolor{darkgreen}{rgb}{0.0,0.45,0.0}
\newcommand{\keywords}[1]{\par\medskip\noindent\textbf{Keywords:}\space\def\and{, }#1}

\title{DeMaVLA: A Vision-Language-Action Foundation Model for Generalizable Deformable Manipulation}

\author[1,*]{Taiyi Su}{}
\author[1,*,\dagger,\ddagger]{Jian Zhu}{}
\author[1,*]{Tianjian Wang}{}
\author[1]{Youzhang He}{}
\author[1,2]{Zitai Huang}{}
\author[1,2]{Jianjun Zhang}{}
\author[1,2]{Chong Ma}{}
\author[1]{Hanyang Wang}{}
\author[1]{Tianjiao Zhang}{}
\author[1]{Munan Yin}{}
\author[1]{Weihao Ding}{}
\author[1,\dagger]{Yi Xu}{}

\affiliation[1]{AIRC, Midea Group}
\affiliation[2]{Tongji University}

\contribution[*]{Equal Contribution}
\contribution[\dagger]{Corresponding Author}
\contribution[\ddagger]{Project Leader}

\abstract{
Real-world household robots require Vision-Language-Action (VLA) foundation models that can acquire reusable manipulation skills across diverse objects, task conditions, and household environments. Deformable-object folding is a representative challenge, requiring robots to handle clothing items from random initial states across varying categories, geometries, materials, and scenes. However, existing VLA systems commonly train separate policies for different object categories, while naively mixed multi-task training often suffers from task interference and degraded performance. To move beyond category-specific folding policies, we introduce \textbf{DeMaVLA}, a VLA foundation model for generalizable \textbf{De}formable \textbf{Ma}nipulation. DeMaVLA adopts a VLM backbone with an action expert and formulates continuous action generation using flow matching. To improve efficiency, the action expert is constructed by pruning every other transformer layer while preserving layer-wise alignment with the VLM backbone, reducing training and inference cost. DeMaVLA is first pre-trained on approximately 5,000 hours of selected real-world dual-arm demonstrations to acquire general manipulation priors. It is then post-trained on mixed folding data that aggregates self-collected demonstrations and corrective trajectories from real-robot failures across multiple folding tasks through a human-in-the-loop Data Aggregation~(DAgger) pipeline. Experiments show that DeMaVLA achieves competitive performance on RoboTwin 2.0 and strong real-world results on our household folding benchmark. These results highlight the value of scalable real-world data, efficient action generation, and corrective learning for general-purpose VLA policies in deformable-object manipulation.

}

\date{\today}
\correspondence{Jian Zhu, Yi Xu}
\checkdata[Project Page]{\url{https://demavla.github.io/}}

\begin{document}
\begingroup
\hypersetup{linkcolor=black}%
\maketitle
\endgroup

\vspace{-4pt}
\begin{figure}[!h]
\centering
\vspace{-10pt}
\includegraphics[width=0.55\textwidth]{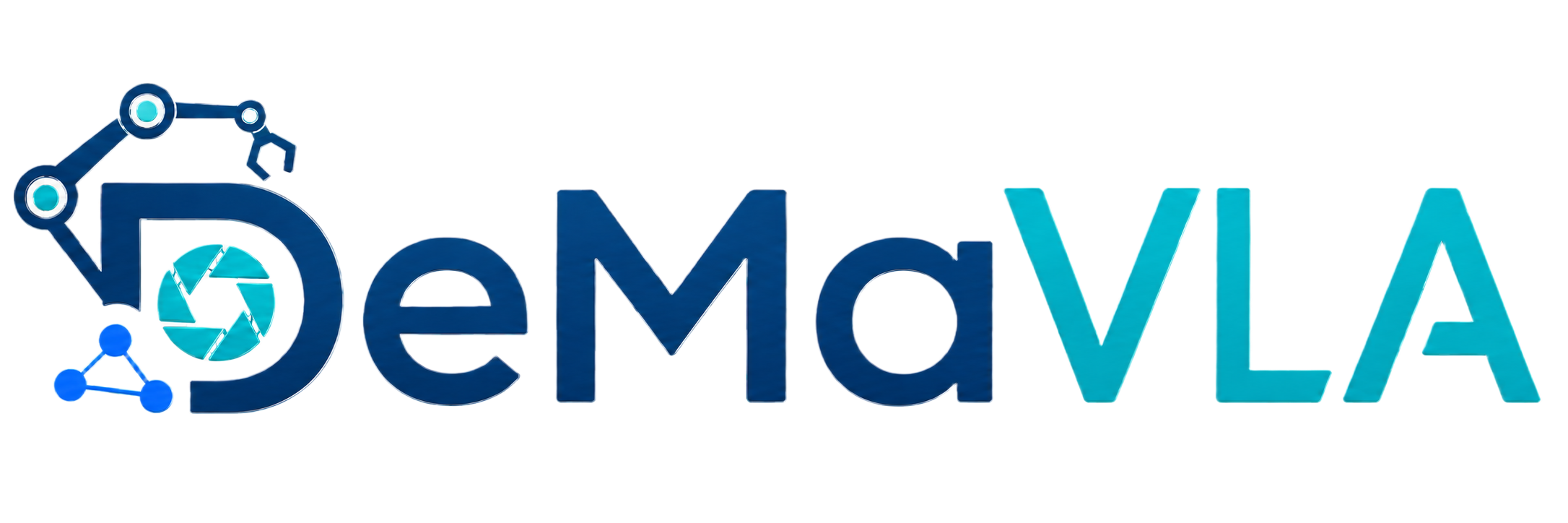}
\end{figure}

\newpage
\tableofcontents
\newpage

\section{Introduction}
\label{sec:intro}

Vision-Language-Action (VLA) foundation models have become a promising route toward general-purpose robot agents~\cite{black2024pi_0,intelligence2025pi_,pertsch2025fast,cheang2025gr,bjorck2025gr00t,zheng2025x,cen2025rynnvla,zhai2025igniting}. By grounding visual observations and language instructions into executable actions, recent systems transfer semantic knowledge from large vision-language models~\cite{beyer2024paligemma,bai2025qwen3} into physical control. Yet real-world household deployment requires more than semantic grounding: a robot must reuse manipulation skills under changing object categories, physical states, task conditions, and scene layouts. Deformable-object folding is a representative example of this challenge. Garment items may appear in random, crumpled, or self-occluded states, and folding shirts, pants, towels, or skirts requires distinct action sequences while still relying on shared physical priors such as grasping corners, aligning fabric, smoothing wrinkles, and completing long-horizon bimanual folds.

Existing VLA-based folding systems typically address this diversity through task-specific adaptation, assigning separate policies to different clothing categories. This strategy is difficult to scale in household environments, where the robot must handle diverse foldable items and random initial states without manually selecting and maintaining many specialized models. A straightforward alternative is to train a single policy on mixed folding data, but naive multi-task training can introduce task interference and degrade performance when different object categories require incompatible intermediate states or motion patterns. These limitations motivate a unified folding policy that can share reusable folding priors across clothing categories while preserving task-specific execution.

To move beyond category-specific folding policies, we introduce \textbf{DeMaVLA}, a VLA foundation model for generalizable \textbf{De}formable \textbf{Ma}nipulation. DeMaVLA is designed as a single-checkpoint policy for multi-category bimanual folding on a dual-arm platform. The model adopts a VLM backbone with an action expert and formulates continuous action generation using flow matching~\cite{lipman2022flow,black2024pi_0}, enabling smooth action prediction for long-horizon dual-arm manipulation. Specifically, our action expert is built from the LLM component of Qwen3-VL, enabling it to share the same transformer architecture as the language backbone. However, directly using the full LLM as an action expert introduces substantial computation, especially because flow-based action generation requires repeated expert forward passes. To retain the aligned architecture while improving efficiency, we prune every other transformer layer in the action expert, reducing training and inference cost without changing the overall VLA interface.

DeMaVLA is trained with both scalable pre-training and failure-driven post-training. It is first pre-trained on approximately 5,000 hours of selected real-world dual-arm demonstrations from public and self-collected datasets to acquire general manipulation priors. It is then post-trained on mixed folding data that combines self-collected demonstrations with corrective trajectories collected through a human-in-the-loop Data Aggregation~(DAgger) pipeline. By rolling out the mixed folding policy on real robots and correcting its failures across different clothing categories, this pipeline directly targets the failure modes of the unified model and substantially improves its multi-category folding capability. We evaluate DeMaVLA on RoboTwin 2.0~\cite{chen2025robotwin} and a real-world household folding benchmark covering multiple clothing categories and random initial states.

Our main contributions are threefold. First, we highlight the importance of implementation-level design in building effective VLA policies, including an LLM-based action expert, skip-layer pruning, flow-matching action generation, and training-time RTC. Second, we demonstrate that scalable real-world pre-training is critical for deformable-object manipulation, as DeMaVLA benefits from approximately 5,000 hours of dual-arm demonstrations and mixed folding data. Third, DeMaVLA achieves strong performance in both simulation and real-world household folding, validating the effectiveness of combining large-scale data, efficient model design, and failure-driven corrective learning.
\section{Related Work}
\label{sec:related}

\subsection{VLA Foundation Models}

Vision-Language-Action (VLA) models have changed robot policy learning by connecting visual perception, language instructions, and action generation in a unified model. Compared with conventional task-specific policies, VLAs provide a scalable interface for instruction following and multi-task manipulation, and have shown strong potential in transferring semantic knowledge into robot control~\cite{zitkovich2023rt,pertsch2025fast,intelligence2025pi,cheang2025gr,bjorck2025gr00t,zheng2025x,cen2025rynnvla,zhai2025igniting}. Most recent VLAs build on pre-trained vision-language models (VLMs), whose image-text pre-training provides useful priors for perception, grounding, and instruction understanding~\cite{bai2025qwen3,beyer2024paligemma}. Robot actions are then learned by adding policy heads or action experts on top of the VLM context~\cite{kim2024openvla,black2024pi_0}.

Scaling embodied data is equally important. Recent foundation policies collect robot trajectories across diverse embodiments, tasks, and environments to improve generalization~\cite{wu2026pragmatic,luo2026being,yang2026abot,jiang2025galaxea}. Related systems further explore cross-embodiment learning, open-world robot datasets, real-time execution, and generalist manipulation policies~\cite{zheng2025x,cen2025rynnvla,team2026gigabrain,lin2026holobrain,zhang2026joyai,cai2026xiaomi,community2026starvla}. For continuous control, flow matching has become a practical formulation because it models multimodal action distributions while generating smooth action chunks~\cite{lipman2022flow,black2024pi_0}.

DeMaVLA follows this VLM- and flow-based VLA paradigm, but focuses on a different deployment goal: learning a unified folding policy across multiple clothing categories and random initial states. Instead of treating each category as an independent downstream adaptation, DeMaVLA learns shared folding priors from large-scale dual-arm data and applies them through a single checkpoint for diverse folding skills.

\begin{figure}[t]
\centering
\includegraphics[width=0.9\linewidth]{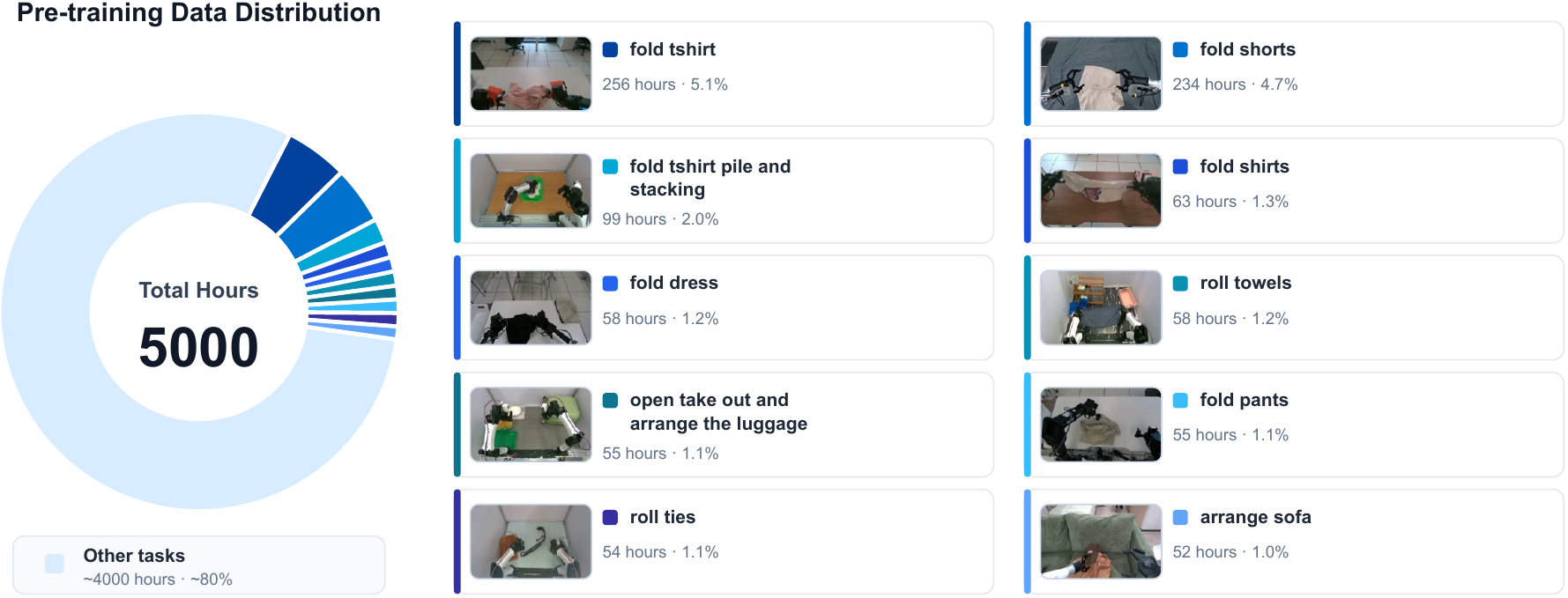}
\caption{Distribution of the DeMaVLA pre-training dataset.}
\label{fig:dataset_distribution}
\end{figure}

\subsection{Interactive Imitation Learning}

Human demonstrations provide direct supervision for complex robot manipulation, making imitation learning a central training strategy for modern VLA policies. In the standard passive setting, expert trajectories are collected first and a policy is trained offline through behavior cloning. This approach is simple and stable, but it suffers from covariate shift: once the learned policy deviates from the demonstration distribution, errors can compound and the robot may fail to recover. Interactive imitation learning addresses this limitation by collecting supervision from the learner's own state distribution~\cite{celemin2022interactive}. 

Data Aggregation (DAgger) formalizes this idea by iteratively executing the current policy, querying expert actions on visited states, and aggregating the newly labeled data into the training set~\cite{ross2011reduction}. Compared with passive behavior cloning, DAgger-style methods improve sample efficiency and robustness because they target states where the learned policy is likely to make mistakes. Human-in-the-loop variants make this idea practical for real robots by allowing a human expert to decide when to intervene. Prior work studies budget-aware querying, teleoperated corrections, recovery data, data augmentation, and human-AI copilot systems~\cite{hoque2021thriftydagger,mandlekar2020human,li2022efficient,hu2025rac,laskin2020reinforcement,li2025gr}. Human-Gated DAgger further lets the operator take over only when needed and release control after correction~\cite{kelly2019hg,wu2025robocopilot,yu2026chi_}.

DeMaVLA adopts this human-gated formulation for multi-category bimanual folding. After initializing the policy from the pre-trained checkpoint, we collect DAgger data by rolling out the policy on real robots and recording corrective human interventions at failure states during mixed folding tasks. This is important for long-horizon deformable-object manipulation, where small errors in grasping, alignment, or intermediate fabric states can compound into failure modes that are rare in static demonstration datasets. By aggregating these corrections into post-training, DeMaVLA directly targets the failure modes of the unified folding policy and improves its robustness across different clothing categories.

\section{The DeMaVLA Model}
\label{sec:method}

\subsection{Pre-training Dataset}
\label{sec:dataset}

Our pre-training dataset contains approximately 5,000 hours of real-world demonstrations from dual-arm humanoid manipulation. Fig.~\ref{fig:dataset_distribution} visualizes the data distribution. The pre-training dataset covers a wide range of real-world bimanual manipulation skills, including household organization, folding, object transfer, industrial handling, and long-horizon coordinated interaction. This broad coverage allows DeMaVLA to acquire general manipulation priors such as reaching, grasping, placement, object rearrangement, and two-arm coordination. At the same time, the folding-oriented subsets provide dense supervision for garment handling, spreading, alignment, and deformable-object manipulation, which are important for the downstream household folding tasks. The combination of broad manipulation data and task-relevant folding data helps DeMaVLA benefit from general-purpose pre-training while remaining compatible with the target ALOHA-style~\cite{fu2024mobile} dual-arm deployment setting.

\begin{figure*}[t]
\centering
\includegraphics[width=\textwidth]{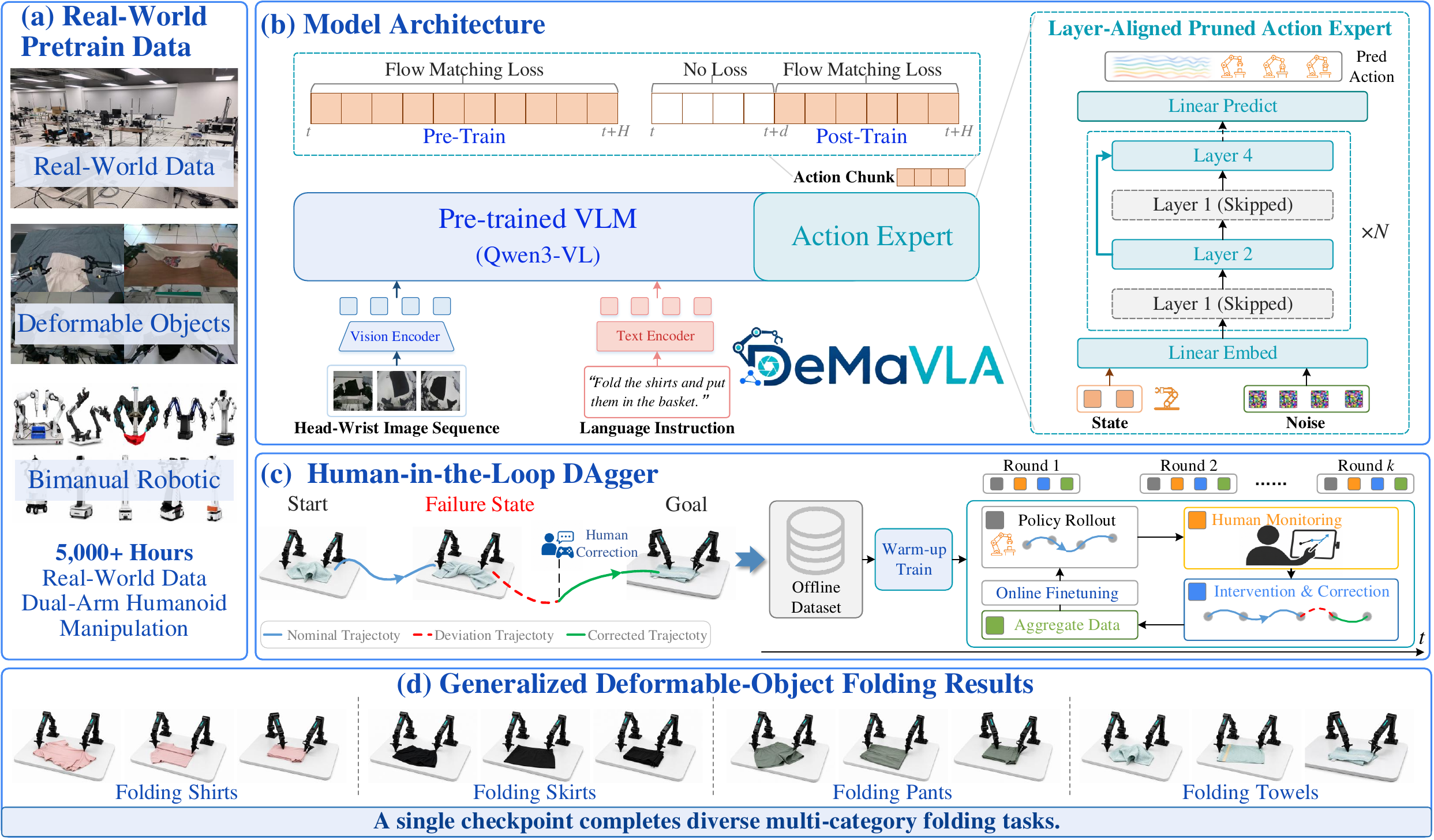}
\caption{\textbf{Overview of DeMaVLA.} DeMaVLA is trained from real-world dual-arm demonstrations and integrates a Qwen3-VL backbone, a layer-aligned pruned action expert, flow-matching action generation, training-time RTC for asynchronous execution, and human-in-the-loop DAgger for corrective real-world learning. A single checkpoint is used to perform diverse multi-category folding tasks.}
\label{fig:method_overview}
\end{figure*}

\subsection{Model Architecture}

DeMaVLA follows a VLM-conditioned action-generation architecture for real-world bimanual manipulation. At each timestep $t$, the model receives multi-view image observations, a language instruction, the robot proprioceptive state, and a noisy action chunk. It predicts the continuous action vector field used for flow-based action generation. The VLM backbone provides visual-language context, while the action expert processes robot-specific inputs and produces continuous bimanual action predictions.

Formally, the observation context is defined as
\begin{equation}
    \mathbf{O}_t = [\mathbf{I}_t^1, \mathbf{I}_t^2, \mathbf{I}_t^3, \ell_t, \mathbf{q}_t],
\end{equation}
where $\mathbf{I}_t^{1,2,3}$ denote the three operational views, $\ell_t$ is the natural language instruction, and $\mathbf{q}_t$ is the robot proprioceptive state. The corresponding action chunk is defined as
\begin{equation}
    \mathbf{A}_t = [\mathbf{a}_t, \mathbf{a}_{t+1}, \dots, \mathbf{a}_{t+H-1}],
\end{equation}
where $H$ is the action horizon and each $\mathbf{a}_t$ is a continuous dual-arm control command.

DeMaVLA adopts Qwen3-VL~\citep{bai2025qwen3} as the VLM backbone. The multi-view image observations and the language instruction are tokenized into visual-language tokens, which provide semantic and spatial context for action generation. Since standard VLM pre-training does not include robot proprioception or continuous action prediction, DeMaVLA introduces an action expert to handle robot-specific tokens. The proprioceptive state $\mathbf{q}_t$ is projected into the transformer embedding space through a linear layer. The noisy action chunk $\mathbf{A}_{t,s}$ is represented as $H$ action tokens, one for each future control step. These tokens are processed by the action expert while being conditioned on the VLM context, and the action expert outputs the vector field used to denoise the action chunk.

\noindent\textbf{Layer-Aligned Pruned Action Expert.}
To align the action expert with the VLM backbone, we build it by reusing the LLM transformer structure of Qwen3-VL. This design preserves a layer-wise correspondence between the visual-language backbone and the action branch, making it straightforward for robot-specific tokens to condition on hierarchical VLM representations. However, directly using the full LLM structure as an action expert is inefficient for flow-based action generation, since the expert must be evaluated repeatedly during denoising.

To retain the aligned architecture while improving efficiency, DeMaVLA constructs a layer-aligned pruned action expert. Let the VLM backbone contain $L$ transformer layers. A full action expert would mirror these $L$ layers for layer-wise conditioning. Instead, DeMaVLA keeps a subset of expert layers and removes the remaining layers from the action branch. For two consecutive retained expert layers, the output of the earlier retained layer is directly passed to the next retained layer. The retained expert layers still correspond to selected layers of the VLM backbone, preserving hierarchical visual-language conditioning while reducing the number of action-expert transformer blocks by approximately half. This produces a lightweight action branch without changing the input-output formulation of the VLA policy.

The resulting architecture combines a Qwen3-VL backbone, a layer-aligned pruned action expert, and flow-matching-based continuous action generation. The parameter counts of the main components are summarized in Table~\ref{tab:model_params}.

\begin{table}[t]
\centering
\caption{Parameter count of DeMaVLA.}
\label{tab:model_params}
\begin{tabular}{lc}
\toprule
\textbf{Module Name} & \textbf{Total Params} \\
\midrule
Vision Encoder & 0.4B \\
Base Language Model & 4.0B \\
Action Expert & 2.2B \\
\midrule
Total Model & 6.6B \\
\bottomrule
\end{tabular}
\end{table}

\subsection{Model Training}

\noindent\textbf{Pre-training.}
During pre-training, DeMaVLA learns a general bimanual manipulation policy from the trajectory mixture described in Sec.~\ref{sec:dataset}. Given the observation context $\mathbf{O}_t$, which contains multi-view images, a language instruction, and robot proprioception, the model predicts a future action chunks. Instead of discretizing robot actions into language tokens, we follow prior VLA policies~\citep{lipman2022flow,black2024pi_0} and model continuous action chunks with conditional flow matching, which is well suited for smooth dual-arm control.

Concretely, the action expert is trained to denoise noisy action chunks conditioned on the VLM context and robot state. This objective encourages the model to learn the conditional action distribution $p(\mathbf{A}_t \mid \mathbf{O}_t)$ while preserving the continuity of the action space. During this stage, the VLM provides visual-language conditioning, and the action expert learns to map noisy action tokens and proprioceptive states into executable dual-arm trajectories. The resulting pre-trained policy provides DeMaVLA with broad manipulation priors before downstream adaptation to real-time folding tasks.

\noindent\textbf{Post-training.}
For real-world deployment, DeMaVLA is post-trained with training-time real-time chunking (RTC)~\citep{black2025training}. Standard action-chunking policies may suffer from inference latency: while the model is generating the next action chunk, the robot must continue executing actions from the previously predicted chunk. Instead of applying inference-time inpainting or guidance~\citep{black2026real}, training-time RTC simulates this latency during post-training and teaches the model to generate future actions conditioned on already committed actions.

Specifically, each action chunk is divided into a committed prefix and a predicted postfix. The committed prefix corresponds to actions that will already be executed before the new prediction becomes available, while the postfix contains the future actions that the model must generate. During post-training, the prefix is kept clean as conditioning context, and flow noise is applied only to the postfix. The training loss is computed only on the postfix actions, encouraging DeMaVLA to produce temporally consistent future trajectories while respecting the committed prefix. At inference time, DeMaVLA conditions on the previously committed actions and predicts the remaining future actions, enabling asynchronous action-chunk execution without additional inpainting or backpropagation-based guidance.

\subsection{Human-in-the-Loop DAgger}

Although large-scale pre-training provides DeMaVLA with broad manipulation priors, real-world multi-category folding still exposes the policy to failure states that are difficult to cover with passive demonstrations alone. Small errors in grasping, cloth alignment, or intermediate folding poses can accumulate over long horizons and lead the robot into states outside the demonstration distribution. To improve robustness under such failures, we adopt a human-gated interactive imitation learning pipeline inspired by HG-DAgger~\citep{kelly2019hg}.

We initialize the policy with behavior cloning on the pre-training and task demonstration data. During interactive teaching, the mixed folding policy is deployed on the real robot and executes folding rollouts across different clothing categories. A human operator monitors the rollout and intervenes when the policy enters an unsafe, unstable, or clearly suboptimal state. Once intervention is triggered, control is fully transferred to the human expert, who corrects the behavior and guides the robot back to a valid task state. After the correction is completed, control is released back to the autonomous policy. This human-gated mechanism avoids requiring the expert to continuously label states under mixed control, and instead collects high-value corrective data around the unified policy's actual failure modes.

Formally, let $\pi_\theta$ denote the current DeMaVLA policy and $\pi_H$ denote the human expert. At timestep $t$, the human operator provides a binary gate $g_t \in \{0,1\}$, where $g_t=1$ indicates human intervention. The executed policy is
\begin{equation}
    \pi_{\mathrm{exec}}(\mathbf{o}_t) =
    g_t \pi_H(\mathbf{o}_t) + (1-g_t)\pi_\theta(\mathbf{o}_t),
\end{equation}
where $\mathbf{o}_t$ denotes the robot observation. In practice, $\mathbf{o}_t$ contains multi-view images, language instruction, and proprioceptive state. We record expert-labeled correction data only during intervention intervals:
\begin{equation}
    \mathcal{D}_{\mathrm{int}} =
    \{(\mathbf{o}_t, \mathbf{a}^H_t) \mid g_t = 1\},
\end{equation}
where $\mathbf{a}^H_t = \pi_H(\mathbf{o}_t)$ is the action provided by the human operator.

After each round of interactive data collection, the intervention data are aggregated into the training set:
\begin{equation}
    \mathcal{D} \leftarrow \mathcal{D} \cup \mathcal{D}_{\mathrm{int}}.
\end{equation}
We then continue post-training DeMaVLA on the updated dataset using the same flow-matching objective. For deployment, we train the final policy on the aggregated dataset collected across all interactive rounds.

Compared with passive imitation learning, this procedure focuses supervision on states where the mixed folding policy is likely to fail. For multi-category folding tasks, such corrective demonstrations are especially valuable because they teach recovery behaviors for misaligned cloth, incomplete grasps, and partially folded configurations. By aggregating these corrections into post-training, DeMaVLA directly targets the failure modes of the unified folding policy and improves its robustness across different clothing categories. As the policy improves, the frequency of human intervention naturally decreases, making the intervention rate a practical signal for measuring whether additional interactive data collection is still needed.

\section{Experiments}
\label{sec:experiment}

\subsection{Comparison on Simulation Benchmark}

To comprehensively evaluate DeMaVLA, we conduct experiments on the RoboTwin 2.0 simulation benchmark~\cite{chen2025robotwin}, which contains 50 bimanual manipulation tasks under both clean and randomized settings. The clean setting uses fixed initial configurations, while the randomized setting varies object poses and scene layouts. We compare DeMaVLA against several representative VLA foundation models including: $\pi_0$~\cite{black2024pi_0}, $\pi_{0.5}$~\cite{intelligence2025pi_}, X-VLA~\cite{zheng2025x}, ABot-M0~\cite{yang2026abot} and LingBot-VLA~\cite{wu2026pragmatic}. All methods are evaluated on the same RoboTwin task suite under clean and randomized settings, and we report the average success rate across the 50 simulation tasks. Table~\ref{tab:model_performance} summarizes the average performance across all tasks, and the full per-task results are reported in Appendix Table~\ref{tab:robotwin_full}.

As shown in Table~\ref{tab:model_performance}, DeMaVLA achieves the best average performance under both settings, reaching 88.42\% in the clean setting and 86.78\% in the randomized setting. The gain under randomized scenes suggests that DeMaVLA preserves strong robustness when object poses and scene layouts vary. Overall, these results indicate that the proposed pre-training recipe and efficient action-generation design provide competitive generalization on diverse bimanual manipulation tasks.

\begin{table}[t]
\centering
\caption{Evaluation on RoboTwin 2.0 simulation benchmark under clean and randomized settings.}
\label{tab:model_performance}
\begin{tabular}{cc cc cc cc cc cc}
\toprule
\multicolumn{2}{c}{$\boldsymbol{\pi_0}$~\cite{black2024pi_0}} &
\multicolumn{2}{c}{$\boldsymbol{\pi_{0.5}}$~\cite{intelligence2025pi_}} &
\multicolumn{2}{c}{\textbf{X-VLA}~\cite{zheng2025x}} &
\multicolumn{2}{c}{\textbf{ABot-M0}~\cite{yang2026abot}} &
\multicolumn{2}{c}{\textbf{LingBot-VLA}~\cite{wu2026pragmatic}} &
\multicolumn{2}{c}{\textbf{DeMaVLA}} \\
\cmidrule(lr){1-2} \cmidrule(lr){3-4} \cmidrule(lr){5-6}
\cmidrule(lr){7-8} \cmidrule(lr){9-10} \cmidrule(lr){11-12}
Clean & Rand. & Clean & Rand. & Clean & Rand. & Clean & Rand. & Clean & Rand. & Clean & Rand. \\
\midrule
65.92 & 58.40 &
82.74 & 76.76 &
72.80 & 72.84 &
80.42 & 81.16 &
86.50 & 85.34 &
\textbf{88.42} & \textbf{86.78} \\
\bottomrule
\end{tabular}
\end{table}

\subsection{Comparison on Real-world Benchmark}
\label{sec:real_world_benchmark}

\noindent\textbf{Task Setup.}
We further evaluate DeMaVLA on a real-world household folding benchmark using an ALOHA-style dual-arm robot. The benchmark contains four representative folding tasks: folding a shirt, folding a skirt, folding pants, and folding a towel. These tasks cover different deformable object geometries, aspect ratios, material properties, and folding routines, making them suitable for evaluating whether a VLA policy can generalize beyond a single object category. Each trial evaluates the complete household folding procedure rather than only the final table-top folding stage. The target item is first randomly dropped into a basket placed beside the table, which differs from settings where the object is already laid flat on the table or neatly stacked in the basket. The robot must pick up the item from the basket, place it on the table, unfold and spread it into a foldable configuration, and then complete the target folding routine. Each task is evaluated over 20 real-world trials using two sets of different items.

\noindent\textbf{Evaluation Protocol.}
A trial is considered successful if the robot completes the full procedure and produces a final folded configuration that satisfies the task-specific completion criteria. A trial is counted as a failure if either (1) the robot does not finish the task within 5 minutes, or (2) the garment falls off the table during execution. We report two metrics for each task: Success Rate (SR), defined as the proportion of successful trials within the 5-minute time limit, and Completion Time, defined as the average task duration. Higher SR indicates better task reliability, while lower completion time indicates more efficient execution. Failed trials are counted as 5 minutes when computing average completion time, so the time metric reflects both execution speed and failure frequency. All methods are evaluated under the same robot hardware, controller frequency, and initialization protocol.

\noindent\textbf{Compared Methods.}
We compare DeMaVLA with a state-of-the-art VLA baseline. Both methods use training-time RTC for asynchronous chunk execution:
\begin{itemize}
    \item $\pi_0$. A strong VLA baseline adapted from the released $\pi_0$ base model. We fine-tune it on our folding tasks with the same training-time RTC setting.
    \item DeMaVLA. Our proposed model, trained with the DeMaVLA architecture, large-scale real-world pre-training data, human-in-the-loop DAgger, and training-time RTC.
\end{itemize}

\noindent\textbf{Single-task Comparison.}
We first compare single-task policies on the shirt folding task, where each model is trained and evaluated only on shirt demonstrations. This setting tests whether the proposed architecture and training recipe provide stronger task-specific folding capability when the task distribution is fixed. As shown in Table~\ref{tab:single_task_avg_results}, single-task DeMaVLA achieves a 100.0\% SR, improving over the single-task $\pi_0$ baseline by 20.0 percentage points. DeMaVLA also slightly reduces the average completion time from 2:13 to 2:04. 

\begin{table}[t]
\centering
\caption{Single-task comparison on the shirt folding task.}
\label{tab:single_task_avg_results}
\begin{tabular}{lcc}
\toprule
\textbf{Training Setting} & \textbf{SR} & \textbf{Time} \\
\midrule
Single-task $\pi_0$ & 80.0\% & 2:13 \\
Single-task DeMaVLA & \textbf{100.0\%} & 2:04 \\
\bottomrule
\end{tabular}
\end{table}

\noindent\textbf{Multi-task Comparison.}
We then evaluate whether one checkpoint can solve multiple folding tasks. In this setting, both $\pi_0$ and DeMaVLA are trained on mixed demonstrations from shirt, skirt, pant, and towel folding, and are evaluated on all four tasks. The mixed folding dataset contains 37 hours of real-robot data in total, including 21.9 hours for folding shirts, 8.4 hours for folding skirts, 3.7 hours for folding pants, and 3.0 hours for folding towels. The DAgger data are collected incrementally following this task order, so later tasks require less additional data as the policy gradually improves and transfers folding priors from earlier tasks.

Table~\ref{tab:real_world_benchmark} reports success rate and average completion time for each task. DeMaVLA achieves a higher average SR than $\pi_0$ across the four tasks, improving from 76.3\% to 92.5\%. The largest gain appears on towel folding, where DeMaVLA reaches 100.0\% SR compared with 55.0\% for $\pi_0$, showing stronger robustness on a highly deformable and visually ambiguous object category. DeMaVLA also improves shirt, skirt, and pant folding SR, while maintaining a lower average completion time across all tasks (2:18 vs. 2:26). These results indicate that DeMaVLA can share folding priors across garment categories and execute them through a single multi-task policy. 

\begin{table}[t]
\centering
\caption{Real-world evaluation on the household folding benchmark.}
\label{tab:real_world_benchmark}
\begin{tabular}{l cc cc cc cc cc}
\toprule
\multirow{2}{*}{\textbf{Method}} &
\multicolumn{2}{c}{\textbf{Shirt}} &
\multicolumn{2}{c}{\textbf{Skirt}} &
\multicolumn{2}{c}{\textbf{Pant}} &
\multicolumn{2}{c}{\textbf{Towel}} &
\multicolumn{2}{c}{\textbf{Average}} \\
\cmidrule(lr){2-3} \cmidrule(lr){4-5} \cmidrule(lr){6-7} \cmidrule(lr){8-9} \cmidrule(lr){10-11}
 & SR & Time & SR & Time & SR & Time & SR & Time & SR & Time \\
\midrule
$\pi_0$ & 90.0\% & 1:55 & 95.0\% & 1:03 & 65.0\% & 3:01 & 55.0\% & 3:44 & 76.3\% & 2:26 \\
DeMaVLA & \textbf{95.0\%} & 2:15 & \textbf{100.0\%} & 1:30 & \textbf{75.0\%} & 3:01 & \textbf{100.0\%} & 2:26 & \textbf{92.5\%} & 2:18 \\
\bottomrule
\end{tabular}
\end{table}

\noindent\textbf{Discussion.}
The real-world benchmark is designed to evaluate two capabilities. First, the comparison with $\pi_0$ tests whether DeMaVLA provides stronger real-world folding performance than a general VLA baseline under the same RTC deployment setting. Second, the single-task and multi-task results examine whether DeMaVLA can use one checkpoint to solve multiple folding tasks while retaining strong performance on the shirt task. Together, these evaluations show that DeMaVLA behaves as a unified fold-anything policy rather than a collection of independently fine-tuned task policies.

\subsection{Pre-training Scaling Analysis}
\label{sec:pretraining_scaling}

We further study how the scale of pre-training data affects downstream real-world folding performance. To isolate the effect of pre-training scale, we compare three DeMaVLA checkpoints trained with different amounts of selected real-world pre-training data: 500 hours, 2,500 hours, and 5,000 hours. All checkpoints are then post-trained with the same shirt-folding data and evaluated under the same real-world protocol. 

\begin{table}[t]
\centering
\caption{Pre-training scaling analysis on the real-world shirt-folding task. All checkpoints are post-trained with the same shirt-folding data.}
\label{tab:pretraining_scaling}
\begin{tabular}{lcc}
\toprule
\textbf{Pre-training Data} & \textbf{SR} & \textbf{Time} \\
\midrule
500 hours & 55.0\% & 3:34 \\
2,500 hours & 70.0\% & 3:21 \\
5,000 hours & \textbf{100.0\%} & 2:04 \\
\bottomrule
\end{tabular}
\end{table}

The results in Table~\ref{tab:pretraining_scaling} show a clear scaling trend. As the pre-training data increases from 500 to 5,000 hours, the SR improves from 55.0\% to 100.0\%, while the average completion time decreases from 3:34 to 2:04. The intermediate 2,500-hour checkpoint also shows consistent improvement, achieving a 70.0\% SR and 3:21 average completion time. These results indicate that scaling real-world pre-training data improves not only downstream task success but also execution efficiency after post-training.

\section{Conclusion}
\label{sec:conclusion}

We introduced DeMaVLA, a VLA foundation model for generalizable deformable manipulation. DeMaVLA moves beyond category-specific folding policies by using a single checkpoint to handle multiple household folding tasks with different garments, initial states, and long-horizon bimanual routines. Its design combines a Qwen3-VL backbone, an LLM-based layer-aligned action expert, skip-layer pruning, flow-matching action generation, and training-time RTC, showing that implementation-level architecture and deployment choices are central to effective real-world VLA policies. To support scalable learning, DeMaVLA is pre-trained on approximately 5,000 hours of selected real-world dual-arm demonstrations and then post-trained with mixed folding data and human-gated DAgger corrections that target policy failure states. Experiments on RoboTwin 2.0 and our real-world household folding benchmark validate this combination of large-scale data, efficient model design, and failure-driven corrective learning, demonstrating strong performance in both simulation and real-world multi-category folding.

\clearpage
\bibliographystyle{plainnat}
\bibliography{main}

\begin{thebibliography}{38}
\providecommand{\natexlab}[1]{#1}
\providecommand{\url}[1]{\texttt{#1}}
\expandafter\ifx\csname urlstyle\endcsname\relax
  \providecommand{\doi}[1]{doi: #1}\else
  \providecommand{\doi}{doi: \begingroup \urlstyle{rm}\Url}\fi

\bibitem[Bai et~al.(2025)Bai, Cai, Chen, Chen, Chen, Cheng, Deng, Ding, Gao,
  Ge, et~al.]{bai2025qwen3}
Shuai Bai, Yuxuan Cai, Ruizhe Chen, Keqin Chen, Xionghui Chen, Zesen Cheng,
  Lianghao Deng, Wei Ding, Chang Gao, Chunjiang Ge, et~al.
\newblock Qwen3-vl technical report.
\newblock \emph{arXiv preprint arXiv:2511.21631}, 2025.

\bibitem[Beyer et~al.(2024)Beyer, Steiner, Pinto, Kolesnikov, Wang, Salz,
  Neumann, Alabdulmohsin, Tschannen, Bugliarello, et~al.]{beyer2024paligemma}
Lucas Beyer, Andreas Steiner, Andr{\'e}~Susano Pinto, Alexander Kolesnikov,
  Xiao Wang, Daniel Salz, Maxim Neumann, Ibrahim Alabdulmohsin, Michael
  Tschannen, Emanuele Bugliarello, et~al.
\newblock Paligemma: A versatile 3b vlm for transfer.
\newblock \emph{arXiv preprint arXiv:2407.07726}, 2024.

\bibitem[Bjorck et~al.(2025)Bjorck, Casta{\~n}eda, Cherniadev, Da, Ding, Fan,
  Fang, Fox, Hu, Huang, et~al.]{bjorck2025gr00t}
Johan Bjorck, Fernando Casta{\~n}eda, Nikita Cherniadev, Xingye Da, Runyu Ding,
  Linxi Fan, Yu~Fang, Dieter Fox, Fengyuan Hu, Spencer Huang, et~al.
\newblock Gr00t n1: An open foundation model for generalist humanoid robots.
\newblock \emph{arXiv preprint arXiv:2503.14734}, 2025.

\bibitem[Black et~al.(2024)Black, Brown, Driess, Esmail, Equi, Finn, Fusai,
  Groom, Hausman, Ichter, et~al.]{black2024pi_0}
Kevin Black, Noah Brown, Danny Driess, Adnan Esmail, Michael Equi, Chelsea
  Finn, Niccolo Fusai, Lachy Groom, Karol Hausman, Brian Ichter, et~al.
\newblock {$\pi_0$}: A vision-language-action flow model for general robot
  control.
\newblock \emph{arXiv preprint arXiv:2410.24164}, 2024.

\bibitem[Black et~al.(2025)Black, Ren, Equi, and Levine]{black2025training}
Kevin Black, Allen~Z Ren, Michael Equi, and Sergey Levine.
\newblock Training-time action conditioning for efficient real-time chunking.
\newblock \emph{arXiv preprint arXiv:2512.05964}, 2025.

\bibitem[Black et~al.(2026)Black, Galliker, and Levine]{black2026real}
Kevin Black, Manuel Galliker, and Sergey Levine.
\newblock Real-time execution of action chunking flow policies.
\newblock \emph{Advances in Neural Information Processing Systems},
  38:\penalty0 33383--33407, 2026.

\bibitem[Cai et~al.(2026)Cai, Guo, He, Jin, Li, Lin, Liu, Liu, Ma, Ma,
  et~al.]{cai2026xiaomi}
Rui Cai, Jun Guo, Xinze He, Piaopiao Jin, Jie Li, Bingxuan Lin, Futeng Liu, Wei
  Liu, Fei Ma, Kun Ma, et~al.
\newblock Xiaomi-robotics-0: An open-sourced vision-language-action model with
  real-time execution.
\newblock \emph{arXiv preprint arXiv:2602.12684}, 2026.

\bibitem[Celemin et~al.(2022)Celemin, P{\'e}rez-Dattari, Chisari, Franzese,
  de~Souza~Rosa, Prakash, Ajanovi{\'c}, Ferraz, Valada, and
  Kober]{celemin2022interactive}
Carlos Celemin, Rodrigo P{\'e}rez-Dattari, Eugenio Chisari, Giovanni Franzese,
  Leandro de~Souza~Rosa, Ravi Prakash, Zlatan Ajanovi{\'c}, Marta Ferraz,
  Abhinav Valada, and Jens Kober.
\newblock Interactive imitation learning in robotics: A survey.
\newblock \emph{Foundations and Trends{\textregistered} in Robotics},
  10\penalty0 (1-2):\penalty0 1--197, 2022.

\bibitem[Cen et~al.(2025)Cen, Huang, Yuan, Li, Yuan, Yu, Jiang, Guo, Li, Luo,
  et~al.]{cen2025rynnvla}
Jun Cen, Siteng Huang, Yuqian Yuan, Kehan Li, Hangjie Yuan, Chaohui Yu, Yuming
  Jiang, Jiayan Guo, Xin Li, Hao Luo, et~al.
\newblock Rynnvla-002: A unified vision-language-action and world model.
\newblock \emph{arXiv preprint arXiv:2511.17502}, 2025.

\bibitem[Cheang et~al.(2025)Cheang, Chen, Cui, Hu, Huang, Kong, Li, Li, Liu,
  Ma, et~al.]{cheang2025gr}
Chilam Cheang, Sijin Chen, Zhongren Cui, Yingdong Hu, Liqun Huang, Tao Kong,
  Hang Li, Yifeng Li, Yuxiao Liu, Xiao Ma, et~al.
\newblock Gr-3 technical report.
\newblock \emph{arXiv preprint arXiv:2507.15493}, 2025.

\bibitem[Chen et~al.(2025)Chen, Chen, Chen, Cai, Liu, Li, Liang, Lin, Ge, Gu,
  et~al.]{chen2025robotwin}
Tianxing Chen, Zanxin Chen, Baijun Chen, Zijian Cai, Yibin Liu, Zixuan Li,
  Qiwei Liang, Xianliang Lin, Yiheng Ge, Zhenyu Gu, et~al.
\newblock Robotwin 2.0: A scalable data generator and benchmark with strong
  domain randomization for robust bimanual robotic manipulation.
\newblock \emph{arXiv preprint arXiv:2506.18088}, 2025.

\bibitem[Community(2026)]{community2026starvla}
StarVLA Community.
\newblock Starvla: A lego-like codebase for vision-language-action model
  developing.
\newblock \emph{arXiv preprint arXiv:2604.05014}, 2026.

\bibitem[Fu et~al.(2024)Fu, Zhao, and Finn]{fu2024mobile}
Zipeng Fu, Tony~Z Zhao, and Chelsea Finn.
\newblock Mobile aloha: Learning bimanual mobile manipulation with low-cost
  whole-body teleoperation.
\newblock \emph{arXiv preprint arXiv:2401.02117}, 2024.

\bibitem[Hoque et~al.(2021)Hoque, Balakrishna, Novoseller, Wilcox, Brown, and
  Goldberg]{hoque2021thriftydagger}
Ryan Hoque, Ashwin Balakrishna, Ellen Novoseller, Albert Wilcox, Daniel~S
  Brown, and Ken Goldberg.
\newblock Thriftydagger: Budget-aware novelty and risk gating for interactive
  imitation learning.
\newblock \emph{arXiv preprint arXiv:2109.08273}, 2021.

\bibitem[Hu et~al.(2025)Hu, Wu, Enock, Li, Kadakia, Erickson, and
  Kumar]{hu2025rac}
Zheyuan Hu, Robyn Wu, Naveen Enock, Jasmine Li, Riya Kadakia, Zackory Erickson,
  and Aviral Kumar.
\newblock Rac: Robot learning for long-horizon tasks by scaling recovery and
  correction.
\newblock \emph{arXiv preprint arXiv:2509.07953}, 2025.

\bibitem[Intelligence et~al.(2025{\natexlab{a}})Intelligence, Amin, Aniceto,
  Balakrishna, Black, Conley, Connors, Darpinian, Dhabalia, DiCarlo,
  et~al.]{intelligence2025pi}
Physical Intelligence, Ali Amin, Raichelle Aniceto, Ashwin Balakrishna, Kevin
  Black, Ken Conley, Grace Connors, James Darpinian, Karan Dhabalia, Jared
  DiCarlo, et~al.
\newblock {$\pi^*_{0.6}$}: a vla that learns from experience.
\newblock \emph{arXiv preprint arXiv:2511.14759}, 2025{\natexlab{a}}.

\bibitem[Intelligence et~al.(2025{\natexlab{b}})Intelligence, Black, Brown,
  Darpinian, Dhabalia, Driess, Esmail, Equi, Finn, Fusai,
  et~al.]{intelligence2025pi_}
Physical Intelligence, Kevin Black, Noah Brown, James Darpinian, Karan
  Dhabalia, Danny Driess, Adnan Esmail, Michael Equi, Chelsea Finn, Niccolo
  Fusai, et~al.
\newblock {$\pi_{0.5}$}: a vision-language-action model with open-world
  generalization.
\newblock \emph{arXiv preprint arXiv:2504.16054}, 2025{\natexlab{b}}.

\bibitem[Jiang et~al.(2025)Jiang, Yuan, Liu, Lu, Cui, Liu, Cheng, Gao, Xu, and
  Zhao]{jiang2025galaxea}
Tao Jiang, Tianyuan Yuan, Yicheng Liu, Chenhao Lu, Jianning Cui, Xiao Liu,
  Shuiqi Cheng, Jiyang Gao, Huazhe Xu, and Hang Zhao.
\newblock Galaxea open-world dataset and g0 dual-system vla model.
\newblock \emph{arXiv preprint arXiv:2509.00576}, 2025.

\bibitem[Kelly et~al.(2019)Kelly, Sidrane, Driggs-Campbell, and
  Kochenderfer]{kelly2019hg}
Michael Kelly, Chelsea Sidrane, Katherine Driggs-Campbell, and Mykel~J
  Kochenderfer.
\newblock Hg-dagger: Interactive imitation learning with human experts.
\newblock In \emph{2019 International Conference on Robotics and Automation
  (ICRA)}, pages 8077--8083. IEEE, 2019.

\bibitem[Kim et~al.(2024)Kim, Pertsch, Karamcheti, Xiao, Balakrishna, Nair,
  Rafailov, Foster, Lam, Sanketi, et~al.]{kim2024openvla}
Moo~Jin Kim, Karl Pertsch, Siddharth Karamcheti, Ted Xiao, Ashwin Balakrishna,
  Suraj Nair, Rafael Rafailov, Ethan Foster, Grace Lam, Pannag Sanketi, et~al.
\newblock Openvla: An open-source vision-language-action model.
\newblock \emph{arXiv preprint arXiv:2406.09246}, 2024.

\bibitem[Laskin et~al.(2020)Laskin, Lee, Stooke, Pinto, Abbeel, and
  Srinivas]{laskin2020reinforcement}
Misha Laskin, Kimin Lee, Adam Stooke, Lerrel Pinto, Pieter Abbeel, and Aravind
  Srinivas.
\newblock Reinforcement learning with augmented data.
\newblock \emph{Advances in neural information processing systems},
  33:\penalty0 19884--19895, 2020.

\bibitem[Li et~al.(2022)Li, Peng, and Zhou]{li2022efficient}
Quanyi Li, Zhenghao Peng, and Bolei Zhou.
\newblock Efficient learning of safe driving policy via human-ai copilot
  optimization.
\newblock \emph{arXiv preprint arXiv:2202.10341}, 2022.

\bibitem[Li et~al.(2025)Li, Ma, Xu, Cui, Cui, Han, Huang, Kong, Liu, Niu,
  et~al.]{li2025gr}
Yunfei Li, Xiao Ma, Jiafeng Xu, Yu~Cui, Zhongren Cui, Zhigang Han, Liqun Huang,
  Tao Kong, Yuxiao Liu, Hao Niu, et~al.
\newblock Gr-rl: Going dexterous and precise for long-horizon robotic
  manipulation.
\newblock \emph{arXiv preprint arXiv:2512.01801}, 2025.

\bibitem[Lin et~al.(2026)Lin, Lin, Du, Xie, Jin, Li, Wu, Wang, Li, Zhao,
  et~al.]{lin2026holobrain}
Xuewu Lin, Tianwei Lin, Yun Du, Hongyu Xie, Yiwei Jin, Jiawei Li, Shijie Wu,
  Qingze Wang, Mengdi Li, Mengao Zhao, et~al.
\newblock Holobrain-0 technical report.
\newblock \emph{arXiv preprint arXiv:2602.12062}, 2026.

\bibitem[Lipman et~al.(2022)Lipman, Chen, Ben-Hamu, Nickel, and
  Le]{lipman2022flow}
Yaron Lipman, Ricky~TQ Chen, Heli Ben-Hamu, Maximilian Nickel, and Matt Le.
\newblock Flow matching for generative modeling.
\newblock \emph{arXiv preprint arXiv:2210.02747}, 2022.

\bibitem[Luo et~al.(2026)Luo, Wang, Zhang, Zheng, Xi, Xu, Xu, Yuan, Zhang,
  Wang, et~al.]{luo2026being}
Hao Luo, Ye~Wang, Wanpeng Zhang, Sipeng Zheng, Ziheng Xi, Chaoyi Xu, Haiweng
  Xu, Haoqi Yuan, Chi Zhang, Yiqing Wang, et~al.
\newblock Being-h0. 5: Scaling human-centric robot learning for
  cross-embodiment generalization.
\newblock \emph{arXiv preprint arXiv:2601.12993}, 2026.

\bibitem[Mandlekar et~al.(2020)Mandlekar, Xu, Mart{\'\i}n-Mart{\'\i}n, Zhu,
  Fei-Fei, and Savarese]{mandlekar2020human}
Ajay Mandlekar, Danfei Xu, Roberto Mart{\'\i}n-Mart{\'\i}n, Yuke Zhu,
  Li~Fei-Fei, and Silvio Savarese.
\newblock Human-in-the-loop imitation learning using remote teleoperation.
\newblock \emph{arXiv preprint arXiv:2012.06733}, 2020.

\bibitem[Pertsch et~al.(2025)Pertsch, Stachowicz, Ichter, Driess, Nair, Vuong,
  Mees, Finn, and Levine]{pertsch2025fast}
Karl Pertsch, Kyle Stachowicz, Brian Ichter, Danny Driess, Suraj Nair, Quan
  Vuong, Oier Mees, Chelsea Finn, and Sergey Levine.
\newblock Fast: Efficient action tokenization for vision-language-action
  models.
\newblock \emph{arXiv preprint arXiv:2501.09747}, 2025.

\bibitem[Ross et~al.(2011)Ross, Gordon, and Bagnell]{ross2011reduction}
St{\'e}phane Ross, Geoffrey Gordon, and Drew Bagnell.
\newblock A reduction of imitation learning and structured prediction to
  no-regret online learning.
\newblock In \emph{Proceedings of the fourteenth international conference on
  artificial intelligence and statistics}, pages 627--635, 2011.

\bibitem[Team et~al.(2026)Team, Wang, Li, Ni, Huang, Zhao, Li, Li, Lv, Liu,
  et~al.]{team2026gigabrain}
GigaBrain Team, Boyuan Wang, Bohan Li, Chaojun Ni, Guan Huang, Guosheng Zhao,
  Hao Li, Jie Li, Jindi Lv, Jingyu Liu, et~al.
\newblock Gigabrain-0.5 m*: a vla that learns from world model-based
  reinforcement learning.
\newblock \emph{arXiv preprint arXiv:2602.12099}, 2026.

\bibitem[Wu et~al.(2025)Wu, Shentu, Liao, Jin, Guo, Sreenath, Lin, and
  Abbeel]{wu2025robocopilot}
Philipp Wu, Yide Shentu, Qiayuan Liao, Ding Jin, Menglong Guo, Koushil
  Sreenath, Xingyu Lin, and Pieter Abbeel.
\newblock Robocopilot: Human-in-the-loop interactive imitation learning for
  robot manipulation.
\newblock \emph{arXiv preprint arXiv:2503.07771}, 2025.

\bibitem[Wu et~al.(2026)Wu, Lu, Wang, Yang, Liu, Wang, Zhu, Sun, Wang, Ma,
  et~al.]{wu2026pragmatic}
Wei Wu, Fan Lu, Yunnan Wang, Shuai Yang, Shi Liu, Fangjing Wang, Qian Zhu,
  He~Sun, Yong Wang, Shuailei Ma, et~al.
\newblock A pragmatic vla foundation model.
\newblock \emph{arXiv preprint arXiv:2601.18692}, 2026.

\bibitem[Yang et~al.(2026)Yang, Zeng, Lin, Chang, Qi, Xiao, Liu, Chen, Chen,
  Huo, et~al.]{yang2026abot}
Yandan Yang, Shuang Zeng, Tong Lin, Xinyuan Chang, Dekang Qi, Junjin Xiao,
  Haoyun Liu, Ronghan Chen, Yuzhi Chen, Dongjie Huo, et~al.
\newblock Abot-m0: Vla foundation model for robotic manipulation with action
  manifold learning.
\newblock \emph{arXiv preprint arXiv:2602.11236}, 2026.

\bibitem[Yu et~al.(2026)Yu, Sima, Jiang, Zhang, Mai, Li, Wang, Chen, Wu, Chen,
  et~al.]{yu2026chi_}
Checheng Yu, Chonghao Sima, Gangcheng Jiang, Hai Zhang, Haoguang Mai, Hongyang
  Li, Huijie Wang, Jin Chen, Kaiyang Wu, Li~Chen, et~al.
\newblock {$\chi_0$}: Resource-aware robust manipulation via taming
  distributional inconsistencies.
\newblock \emph{arXiv preprint arXiv:2602.09021}, 2026.

\bibitem[Zhai et~al.(2025)Zhai, Liu, Fang, Cai, Ma, Yin, Wang, Zhou, Wang, Shi,
  et~al.]{zhai2025igniting}
Andy Zhai, Brae Liu, Bruno Fang, Chalse Cai, Ellie Ma, Ethan Yin, Hao Wang,
  Hugo Zhou, James Wang, Lights Shi, et~al.
\newblock Igniting vlms toward the embodied space.
\newblock \emph{arXiv preprint arXiv:2509.11766}, 2025.

\bibitem[Zhang et~al.(2026)Zhang, Yuan, Chi, Liu, Li, Hu, Zhang, Nie, Wei,
  Chen, et~al.]{zhang2026joyai}
Tianle Zhang, Zhihao Yuan, Dafeng Chi, Peidong Liu, Dongwei Li, Kejun Hu, Likui
  Zhang, Junnan Nie, Ziming Wei, Zengjue Chen, et~al.
\newblock Joyai-ra 0.1: A foundation model for robotic autonomy.
\newblock \emph{arXiv preprint arXiv:2604.20100}, 2026.

\bibitem[Zheng et~al.(2025)Zheng, Li, Wang, Liu, Kang, Feng, Zheng, Zou, Chen,
  Zeng, et~al.]{zheng2025x}
Jinliang Zheng, Jianxiong Li, Zhihao Wang, Dongxiu Liu, Xirui Kang, Yuchun
  Feng, Yinan Zheng, Jiayin Zou, Yilun Chen, Jia Zeng, et~al.
\newblock X-vla: Soft-prompted transformer as scalable cross-embodiment
  vision-language-action model.
\newblock \emph{arXiv preprint arXiv:2510.10274}, 2025.

\bibitem[Zitkovich et~al.(2023)Zitkovich, Yu, Xu, Xu, Xiao, Xia, Wu, Wohlhart,
  Welker, Wahid, et~al.]{zitkovich2023rt}
Brianna Zitkovich, Tianhe Yu, Sichun Xu, Peng Xu, Ted Xiao, Fei Xia, Jialin Wu,
  Paul Wohlhart, Stefan Welker, Ayzaan Wahid, et~al.
\newblock Rt-2: Vision-language-action models transfer web knowledge to robotic
  control.
\newblock In \emph{Conference on Robot Learning}, pages 2165--2183. PMLR, 2023.

\end{thebibliography}

\clearpage

\beginappendix

\setcounter{figure}{0}
\setcounter{table}{0}
\renewcommand{\thefigure}{A.\arabic{figure}}
\renewcommand{\thetable}{A.\arabic{table}}



\begin{table}[!htbp]
\centering
\caption{Evaluation on RoboTwin 2.0 Simulation Benchmark.}
\label{tab:robotwin_full}
\setlength{\tabcolsep}{3.6pt}
\resizebox{0.85\textwidth}{!}{%
\begin{tabular}{l cc cc cc cc cc}
\toprule
\multirow{2}{*}{\textbf{Simulation Task}} &
\multicolumn{2}{c}{$\pi_0$} & \multicolumn{2}{c}{$\pi_{0.5}$} & \multicolumn{2}{c}{X-VLA} & \multicolumn{2}{c}{LingBot-VLA} & \multicolumn{2}{c}{DeMaVLA} \\
\cmidrule(lr){2-3} \cmidrule(lr){4-5} \cmidrule(lr){6-7} \cmidrule(lr){8-9} \cmidrule(lr){10-11}
 & Clean & Rand. & Clean & Rand. & Clean & Rand. & Clean & Rand. & Clean & Rand. \\
\midrule
\textit{Adjust Bottle} & 99\% & 95\% & 100\% & 99\% & 100\% & 99\% & 100\% & 100\% & 99\% & 100\% \\
\textit{Beat Block Hammer} & 79\% & 84\% & 96\% & 93\% & 92\% & 88\% & 87\% & 91\% & 79\% & 85\% \\
\textit{Blocks Ranking RGB} & 80\% & 63\% & 92\% & 85\% & 83\% & 83\% & 92\% & 91\% & 95\% & 95\% \\
\textit{Blocks Ranking Size} & 14\% & 5\% & 49\% & 26\% & 67\% & 74\% & 66\% & 73\% & 72\% & 68\% \\
\textit{Click Alarmclock} & 77\% & 68\% & 98\% & 89\% & 99\% & 99\% & 93\% & 26\% & 98\% & 100\% \\
\textit{Click Bell} & 71\% & 48\% & 99\% & 66\% & 100\% & 100\% & 32\% & 19\% & 96\% & 98\% \\
\textit{Dump Bin Bigbin} & 88\% & 83\% & 92\% & 97\% & 79\% & 77\% & 97\% & 92\% & 91\% & 94\% \\
\textit{Grab Roller} & 98\% & 94\% & 100\% & 100\% & 100\% & 100\% & 100\% & 99\% & 100\% & 100\% \\
\textit{Handover Block} & 47\% & 31\% & 66\% & 57\% & 73\% & 37\% & 80\% & 83\% & 93\% & 83\% \\
\textit{Handover Mic} & 97\% & 97\% & 98\% & 97\% & 0\% & 0\% & 94\% & 98\% & 95\% & 96\% \\
\textit{Hanging Mug} & 14\% & 11\% & 18\% & 17\% & 23\% & 27\% & 32\% & 27\% & 46\% & 35\% \\
\textit{Lift Pot} & 80\% & 72\% & 96\% & 85\% & 99\% & 100\% & 100\% & 99\% & 97\% & 94\% \\
\textit{Move Can Pot} & 68\% & 48\% & 51\% & 55\% & 89\% & 86\% & 79\% & 84\% & 93\% & 79\% \\
\textit{Move Pillbottle Pad} & 67\% & 46\% & 84\% & 61\% & 73\% & 71\% & 93\% & 94\% & 86\% & 86\% \\
\textit{Move Playingcard Away} & 74\% & 65\% & 96\% & 84\% & 93\% & 98\% & 96\% & 99\% & 94\% & 90\% \\
\textit{Move Stapler Pad} & 41\% & 24\% & 56\% & 42\% & 78\% & 73\% & 74\% & 49\% & 83\% & 76\% \\
\textit{Open Laptop} & 71\% & 81\% & 90\% & 96\% & 93\% & 100\% & 96\% & 96\% & 98\% & 100\% \\
\textit{Open Microwave} & 4\% & 32\% & 34\% & 77\% & 79\% & 71\% & 91\% & 75\% & 83\% & 72\% \\
\textit{Pick Diverse Bottles} & 69\% & 31\% & 81\% & 71\% & 58\% & 36\% & 79\% & 86\% & 58\% & 75\% \\
\textit{Pick Dual Bottles} & 59\% & 37\% & 93\% & 63\% & 47\% & 36\% & 82\% & 95\% & 89\% & 75\% \\
\textit{Place A2B Left} & 43\% & 47\% & 87\% & 82\% & 48\% & 49\% & 86\% & 83\% & 92\% & 95\% \\
\textit{Place A2B Right} & 39\% & 34\% & 87\% & 84\% & 36\% & 36\% & 74\% & 77\% & 90\% & 87\% \\
\textit{Place Bread Basket} & 62\% & 46\% & 77\% & 64\% & 81\% & 71\% & 92\% & 93\% & 83\% & 84\% \\
\textit{Place Bread Skillet} & 66\% & 49\% & 85\% & 66\% & 77\% & 67\% & 90\% & 89\% & 94\% & 83\% \\
\textit{Place Burger Fries} & 81\% & 76\% & 94\% & 87\% & 94\% & 94\% & 95\% & 96\% & 96\% & 95\% \\
\textit{Place Can Basket} & 55\% & 46\% & 62\% & 62\% & 49\% & 52\% & 68\% & 78\% & 87\% & 77\% \\
\textit{Place Cans Plasticbox} & 63\% & 45\% & 94\% & 84\% & 97\% & 98\% & 97\% & 100\% & 89\% & 92\% \\
\textit{Place Container Plate} & 97\% & 92\% & 99\% & 95\% & 97\% & 95\% & 99\% & 99\% & 96\% & 98\% \\
\textit{Place Dual Shoes} & 59\% & 51\% & 75\% & 75\% & 79\% & 88\% & 80\% & 83\% & 96\% & 94\% \\
\textit{Place Empty Cup} & 91\% & 85\% & 100\% & 99\% & 100\% & 98\% & 100\% & 100\% & 99\% & 99\% \\
\textit{Place Fan} & 66\% & 71\% & 87\% & 85\% & 80\% & 75\% & 91\% & 79\% & 94\% & 91\% \\
\textit{Place Mouse Pad} & 20\% & 20\% & 60\% & 39\% & 70\% & 70\% & 82\% & 78\% & 75\% & 78\% \\
\textit{Place Object Basket} & 67\% & 70\% & 80\% & 76\% & 44\% & 39\% & 90\% & 91\% & 84\% & 66\% \\
\textit{Place Object Scale} & 57\% & 52\% & 86\% & 80\% & 52\% & 74\% & 84\% & 90\% & 90\% & 89\% \\
\textit{Place Object Stand} & 82\% & 68\% & 91\% & 85\% & 86\% & 88\% & 97\% & 93\% & 93\% & 92\% \\
\textit{Place Phone Stand} & 49\% & 53\% & 81\% & 81\% & 88\% & 87\% & 92\% & 93\% & 95\% & 90\% \\
\textit{Place Shoe} & 76\% & 76\% & 92\% & 93\% & 96\% & 95\% & 99\% & 94\% & 100\% & 100\% \\
\textit{Press Stapler} & 44\% & 37\% & 87\% & 83\% & 92\% & 98\% & 90\% & 88\% & 96\% & 97\% \\
\textit{Put Bottles Dustbin} & 65\% & 56\% & 84\% & 79\% & 74\% & 77\% & 88\% & 92\% & 88\% & 85\% \\
\textit{Put Object Cabinet} & 73\% & 60\% & 80\% & 79\% & 46\% & 48\% & 92\% & 86\% & 92\% & 85\% \\
\textit{Rotate QRcode} & 74\% & 70\% & 89\% & 87\% & 34\% & 33\% & 93\% & 84\% & 95\% & 86\% \\
\textit{Scan Object} & 55\% & 42\% & 72\% & 65\% & 14\% & 36\% & 91\% & 97\% & 79\% & 83\% \\
\textit{Shake Bottle Horizontally} & 98\% & 92\% & 99\% & 99\% & 100\% & 100\% & 100\% & 100\% & 100\% & 99\% \\
\textit{Shake Bottle} & 94\% & 91\% & 99\% & 97\% & 99\% & 100\% & 99\% & 100\% & 100\% & 100\% \\
\textit{Stack Blocks Three} & 72\% & 52\% & 91\% & 76\% & 6\% & 10\% & 92\% & 99\% & 97\% & 95\% \\
\textit{Stack Blocks Two} & 93\% & 79\% & 97\% & 100\% & 92\% & 87\% & 100\% & 100\% & 100\% & 98\% \\
\textit{Stack Bowls Three} & 77\% & 75\% & 77\% & 71\% & 76\% & 86\% & 72\% & 83\% & 83\% & 80\% \\
\textit{Stack Bowls Two} & 94\% & 95\% & 95\% & 96\% & 96\% & 93\% & 92\% & 95\% & 98\% & 96\% \\
\textit{Stamp Seal} & 46\% & 33\% & 79\% & 55\% & 76\% & 82\% & 76\% & 86\% & 77\% & 83\% \\
\textit{Turn Switch} & 41\% & 42\% & 62\% & 54\% & 40\% & 61\% & 61\% & 65\% & 18\% & 41\% \\
\midrule
\textbf{Average (\%)} & 65.92 & 58.40 & 82.74 & 76.76 & 72.80 & 72.84 & 86.50 & 85.34 & \textbf{88.42} & \textbf{86.78} \\
\bottomrule
\end{tabular}%
}
\vspace{-5pt}
\end{table}

\end{document}